\documentclass[conference]{IEEEtran}
\usepackage{cite}
\usepackage{amsmath,amssymb,amsfonts}
\usepackage{algorithm}
\usepackage{algpseudocode}
\usepackage{graphicx}
\usepackage{textcomp}
\usepackage{xcolor}
\usepackage{graphics}
\usepackage{MnSymbol}

\newcommand{\always}{\square}
\newcommand{\eventually}{\lozenge}
\newcommand{\until}{\mathcal{U}_I}

\newcommand{\ew}{\boxbox_{\ra}}
\newcommand{\sw}{\diamonddiamond_{\ra}}
\newcommand{\ag}{\mathcal{A}_{\ra}}
\newcommand{\ct}{\mathcal{C}_{\ra}}

\newcommand{\op}{\mathrm{op}}

\newcommand{\avg}{\mathrm{avg}}

\newcommand{\nb}{L_{\ra}}
\newcommand{\nbx}{\alpha_{\ra}^x(\omega, t, l)}

\newcommand{\ra}{\mathcal{D}}

\definecolor{brilliantlavender}{rgb}{0.96, 0.73, 1.0}
\definecolor{blond}{rgb}{0.98, 0.94, 0.75}
\definecolor{celadon}{rgb}{0.67, 0.88, 0.69}
\definecolor{columbiablue}{rgb}{0.61, 0.87, 1.0}
\definecolor{lavenderblush}{rgb}{1.0, 0.94, 0.96}
\definecolor{electriclavender}{rgb}{0.96, 0.73, 1.0}

\begin{document}

\title{CitySpec: An Intelligent Assistant System for Requirement Specification in Smart Cities}

% \and

\author{
	\IEEEauthorblockN{Zirong Chen\IEEEauthorrefmark{1}, Isaac Li\IEEEauthorrefmark{2}, Haoxiang Zhang\IEEEauthorrefmark{3}, Sarah Preum\IEEEauthorrefmark{4}, John A. Stankovic\IEEEauthorrefmark{2}, Meiyi Ma\IEEEauthorrefmark{1}  \\}
	\IEEEauthorblockA{\IEEEauthorrefmark{1}
 Vanderbilt University  \IEEEauthorrefmark{2} University of Virginia
 \IEEEauthorrefmark{3}  Columbia University
 \IEEEauthorrefmark{4} Dartmouth College
 \\ \{zirong.chen, meiyi.ma\}@vanderbilt.edu \{il5fq, stankovic\}@virginia.edu sarah.masud.preum@dartmouth.edu
		}
	
}

% 	\IEEEauthorblockA{\IEEEauthorrefmark{1}
% 		Department of Computer Science, Vanderbilt University, Nashville, TN 37235, USA \{1, 4\}@abc.com
% 		}

\maketitle

\begin{abstract}
An increasing number of monitoring systems have been developed in smart cities to ensure that a city’s real-time operations satisfy safety and performance requirements. However, many existing city requirements are written in English with missing, inaccurate, or ambiguous information. There is a high demand for assisting city policy makers in converting human-specified requirements to machine-understandable formal specifications for monitoring systems. To tackle this limitation, we build \textit{CitySpec}, the first intelligent assistant system for requirement specification in smart cities. To create CitySpec, we first collect over 1,500 real-world city requirements across different domains from over 100 cities and extract city-specific knowledge to generate a dataset of city vocabulary with 3,061 words. We also build a translation model and enhance it through requirement synthesis and develop a novel online learning framework with validation under uncertainty. The evaluation results on real-world city requirements show that CitySpec increases the sentence-level accuracy of requirement specification from 59.02\% to 86.64\%, and has strong adaptability to a new city and a new domain (e.g., F1 score for requirements in Seattle increases from 77.6\% to 93.75\% with online learning).

\end{abstract}
% The present-day city policymakers want to put into place a monitoring system for sensor-collected data that will help them make informed decisions. As smart city development gets more and more popular, there is a growing need to interpret, analyze, and monitor sensor-collected data. This need comes along with greater costs in terms of allocating resources to the necessary parties. As it would be inefficient to thoroughly train the city policymakers to learn everything it takes to master the complicated the machine-compatible input languages, researchers need to focus on refining the machines to understand the intentions of the policymakers. However, the monitoring system only takes machine-compatible input languages as input, which will bring difficulties during real-world deployment. In this work, we propose a novel method with a user-friendly interface to convert natural languages to the machine-compatible input languages, which is formal specifications in this work. The results of the experiments show that this approach (1) handles inaccurate, missing, and ambiguous inputs; (2) composes cognitive assistance by launching human-in-the-loop correction and clarification; (3) continuously learning online and offline from knowledge injection. Due to the limited data access, a controllable augmentation method is introduced and experimented with. 


\section{Introduction}
\label{sec:introduction}

\begin{figure}[t]
    \centering
    \includegraphics[width=0.48\textwidth]{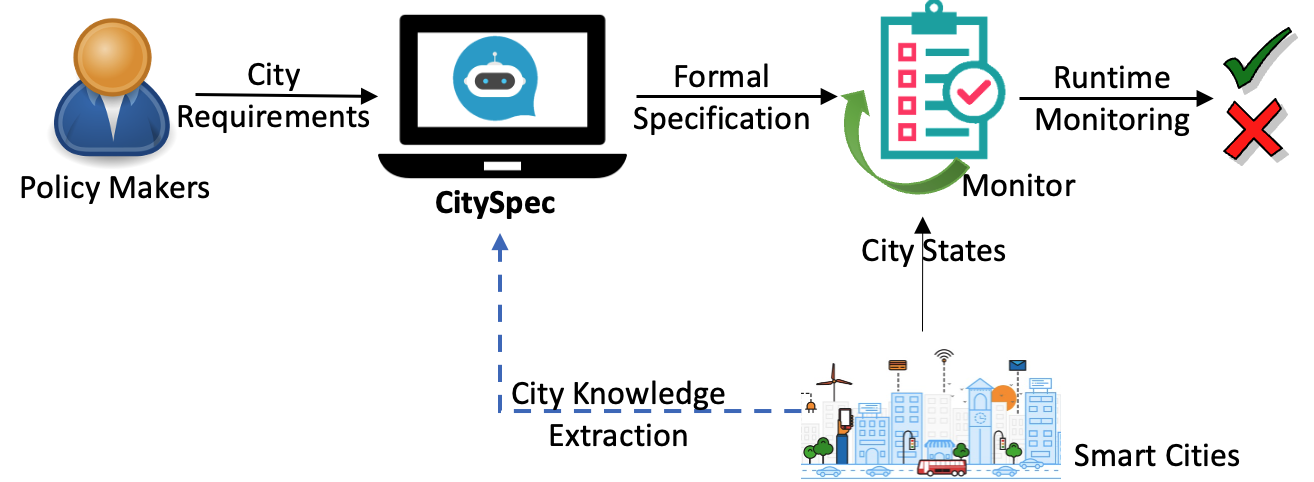}
    \caption{CitySpec in Smart Cities}
    \label{fig:CitySpec_City}
    \vspace{-0.8cm}
\end{figure}

With the increasing demand for safety guarantees in smart cities, significant research efforts have been spent toward how to ensure that a city’s real-time operations satisfy safety and performance requirements~\cite{ma2021toward}. Monitoring systems, such as SaSTL runtime monitoring~\cite{ma2021novel}, CityResolver~\cite{ma2018cityresolver}, and STL-U predictive monitoring~\cite{ma2021predictive}, have been developed in smart cities. Figure \ref{fig:CitySpec_City} shows a general framework of monitoring systems in smart cities. These systems are designed to execute in city centers and to support decision-making based on the verification results of real-time sensing data about city-states (such as traffic and air pollution). If the monitor detects a requirement violation, the city operators can take actions to change the states, such as improving air quality, sending alarms to police, calling an ambulance, etc.

The monitor systems have two important inputs, i.e., the real-time data streams and formal specified requirements. Despite that extensive research efforts have been spent toward improving the expressiveness of specification languages and efficiency of the monitoring algorithms, the research challenge of how to convert human-specified requirements to machine-understandable formal specifications has received only scant attention. 
Moreover, our study (see Section \ref{sec:motivation}) on over 1,500 real-world city requirements across different domains shows that,  first, existing city requirements are often defined with missing information or ambiguous description, e.g., no location information, using words like nearby, or close to. They are not precise enough to be converted to a formal specification or monitored in a city directly without clarifications by policy makers. Secondly,
the language difference between English specified requirements and formalized specifications is significant. Without expertise in formal languages, it is extremely difficult or impossible for policy makers to write or convert their requirements to formal specifications. Therefore, there is an urgent demand for an intelligent system to support policy makers for requirement specifications in smart cities.   

Despite the prevalence of developing models to translate natural language to machine languages in various domains, such as Bash commands~\cite{fu2021transformer}, Seq2SQL~\cite{zhong2017seq2sql}, and Python~\cite{chen2021evaluating}, it is very \textbf{challenging} to develop such an intelligent system for requirement specification in smart cities for the following reasons. First, unlike the above translation tasks with thousands or even millions of samples in a dataset, there is barely any requirement specification data. 
As a result, traditional language models are not sufficient to be applied directly. Moreover, the requirements usually contain city domain-specific descriptions and patterns that existing pre-trained embeddings like BERT or GloVE cannot handle effectively. 
Furthermore, requirements from different domains and cities vary significantly and evolve over time, thus building a system that can adapt to new domains at runtime is an open research question. Good adaptability can increase user experience (e.g., policy makers do not have to clarify new terms repeatedly), while one of the major challenges is validating and filtering the new knowledge and avoiding adversarial examples online.      

% intelligent
% adaptive; validation  

% Missing, incorrect, ambiguous requirements 
% Small Data + Very domain specific – (almost) no existing work 
% Adapt to new knowledge from different cities, different domains/departments, different users, + overtime  
% Online learning, how to validate new knowledge? 

In this paper, we target the above technical challenges, and describe \textit{CitySpec}, an intelligent assistant system for requirement specification in smart cities. To the best of our knowledge, it is the first specification system helping city policy makers specify and translate their requirements into formal specifications automatically. As shown in Figure \ref{fig:CitySpec_City}, CitySpec is designed to bridge the gap between city policy makers and monitoring systems. It enables policy makers to define their requirements by detecting missing, inaccurate, or ambiguous information through an intelligent assistant interface. To effectively train the translation model using a small amount of city requirement data, CitySpec extracts city knowledge and enhances the learning process through requirement synthesis. CitySpec can easily adapt to a new city or domain through online learning and validation.

\textbf{Contributions}. We summarize the major contributions of this paper as follows: 
\begin{itemize}
    \item We collect and annotate over 1,500 real-world city requirements from over 100 cities across different domains. We extract city-specific knowledge and build a dataset of city vocabulary with 3,061 words in 5 categories. 
    % \footnote{Both datasets will be available with the paper.}. 
    \item We create an intelligent assistant system for requirement specification in smart cities. 
    In the system, we build a translation model and enhance it through requirement synthesis, and 
     develop a novel online learning framework with validation under uncertainty. 
    \item We evaluate CitySpec extensively using real-world city requirements. The evaluation results show that CitySpec is effective on supporting policy makers accurately writing and refining their requirements. It increases the sentence level accuracy of requirement specification from 59.02\% to 86.64\% through city knowledge injection. It shows strong adaptability (user experience) to a new city (e.g., F1 score in Seattle from 77.6\% to 93.75\%) and a new domain (e.g., F1 score in security domain from 62.93\% to 93.95\%). 
\end{itemize}

% (1) the detection and correction of missing, inaccurate or ambiguous information in input requirements; (2) the automatic conversion from natural languages to formal specifications; (3) the Natural Language Processing (NLP) features-based augmentation approach to grant the training process more efficiency on relatively small datasets.

% Cognitive assistant system with three major layers
% Front; middle - online; bottom 
% Help users formulate the requirements – conversation 
% - novel 
% 1000 + quantitative-specified requirements  
% 500+ other requirements 
% City vocabulary: 2500+ words (over 5 category)
% Publish the dataset 
% Novel technique for online learning with validation –
% other than SC. General
% E.g., Hospitals -- 
% Evaluation: 96% token-level accuracy, 85% sentence-level accuracy 
% Compare to the baselines 

\textbf{Paper organization}: In the rest of the paper, we describe the motivating study of city requirement specification  in Section \ref{sec:motivation}, provide an overview of CitySpec in Section \ref{sec:overview}, and present the technical details in Section \ref{sec:method}. We then present the evaluation results in Section \ref{sec:eval}, discuss the related work in Section \ref{sec:related} and draw conclusions in Section \ref{sec:summary}.

% Different from traditional NLP translation tasks, there is a very limited number of city requirements with corresponding formal specification. 
% At the training time (as indicated by the green arrows in Figure \ref{fig:Overview}), CitySpec extracts city knowledge from real-world city requirements and synthesize a large number of requirements to enhance the learning process. 

% However, as an under-exploited domain, there is a very limited number of well-defined requirements. Annotation of formal specification requires specialties in formal methods and is extremely time-consuming. 
%
\section{Motivating Study}
\label{sec:motivation}

\begin{table*}[t]
\caption{Comparison between English requirements and formal specifications}
\scriptsize
\centering{%
\begin{tabular}{|c|p{7.8cm}|p{7.5cm}|{c}|}
\hline
\textbf{ID} & \textbf{English Requirement }                                                                                            & \textbf{Formal Specification }                                                                                                                       & \textbf{DLD}  \\ [0.2ex]\hline\hline
\textbf{1}  & Sliding glass doors shall have an  air infiltration rate of no more than 0.3 cfm per square foot.               & $\mathsf{Everywhere}_\mathsf{{Sliding ~ glass ~ doors}} ~ \mathsf{Always}_{[0, +\infty)} ~ \mathsf{air~infiltration~rate} ~ \leq 0.3 ~ \mathsf{cfm/foot}^{2}$ & 59  \\ \hline
\textbf{2}  & The operation of a Golf Cart upon a Golf Cart Path shall be restricted to a maximum speed of 15 miles per hour from 8:00 to 16:00. & $\mathsf{Everywhere}_\mathsf{Golf~ Cart ~Path}\mathsf{Always}_{[8,16]}\mathsf{Golf ~Cart ~speed} < 15$ $~\mathsf{miles/hour}$       & 67  \\ \hline
\textbf{3}  & Up to four vending vehicles may dispense merchandise in any given city block at any time.                       & $\mathsf{Everywhere}_\mathsf{city ~block}\mathsf{Always}_{[0, +\infty)}\mathsf{vending ~vehicles}  \leq 4$                              & 75  \\ \hline
\end{tabular}
}
\label{tab:eng_formal}
\vspace{-0.3cm}
\end{table*}

In this section, we study real-world city requirements
and their formal specification as motivating examples to discuss the demand and challenges of developing an intelligent assistant system for requirement specification in smart cities. 
We collect and annotate over 1,500 real-world city requirements across different domains (e.g., transportation, environment, security, public safety, indoor environments, etc.) from over 100 cities. 
We make the following observations from the analysis of the requirement dataset.

\textit{Existing city requirements are often defined with missing information or ambiguous description.} In \cite{ma2021novel}, the authors define essential elements for monitoring a city requirement.   
Within the 1500 requirements, many requirements have one or more missing elements. For example, 27.6\% of the requirements do not have location information, 29.1\% of the requirements do not have a proper quantifier, and 90\% of the requirements do not have or only have a default time (e.g., always) defined. Additionally, requirements often have ambiguous descriptions that are difficult to be noticed by policy makers. For example, a location is specified as ``nearby'' or ``close to''. As a result, it is very difficult or impossible for the monitoring system to monitor these requirements properly. It indicates a high demand for an intelligent assistant system to support the policy makers to refine the requirements.

% ratio of missing key1: 0.00558659217877095
% ratio of missing key2: 0.2918994413407821
% ratio of missing lockey: 0.27583798882681565
% ratio of missing number: 0.5747206703910615
% ratio of missing time: 0.9392458100558659

\textit{The language difference between English specified requirements and their formal specifications is significant.}
In Table \ref{tab:eng_formal}, we give three examples of city requirements in English, their formal specification in SaSTL, and the Damerau–Levenshtein Distance (DLD)~\cite{damerau1964technique} between each pair of requirements. DLD measures the edit distance between two sequences. 
It shows that 
natural languages are different from machine-compatible input languages. Formal specifications usually consist of mathematical symbols, which makes the conversion even more difficult. 
As shown in Table \ref{tab:eng_formal}, the average DLD from English requirements to formal specifications is 67, which means that it requires an average of 67 edits. 
As a reference, the average DLD brought by translating these three English requirements to Latin is 64.67. It indicates that the conversion from English requirements to formal specifications even requires more edits than the translation of these requirements from English to Latin. In general, building a translator from English to Latin would require millions of samples. However, as an under-exploited domain, there is a very limited number of well-defined requirements. Moreover, annotation of formal specifications requires specialties in formal methods and is extremely time-consuming. It presents major challenges for building such a translation model.

\section{System Overview}
\label{sec:overview}

\begin{figure*}[t]
    \centering
    \includegraphics[width=0.85\textwidth]{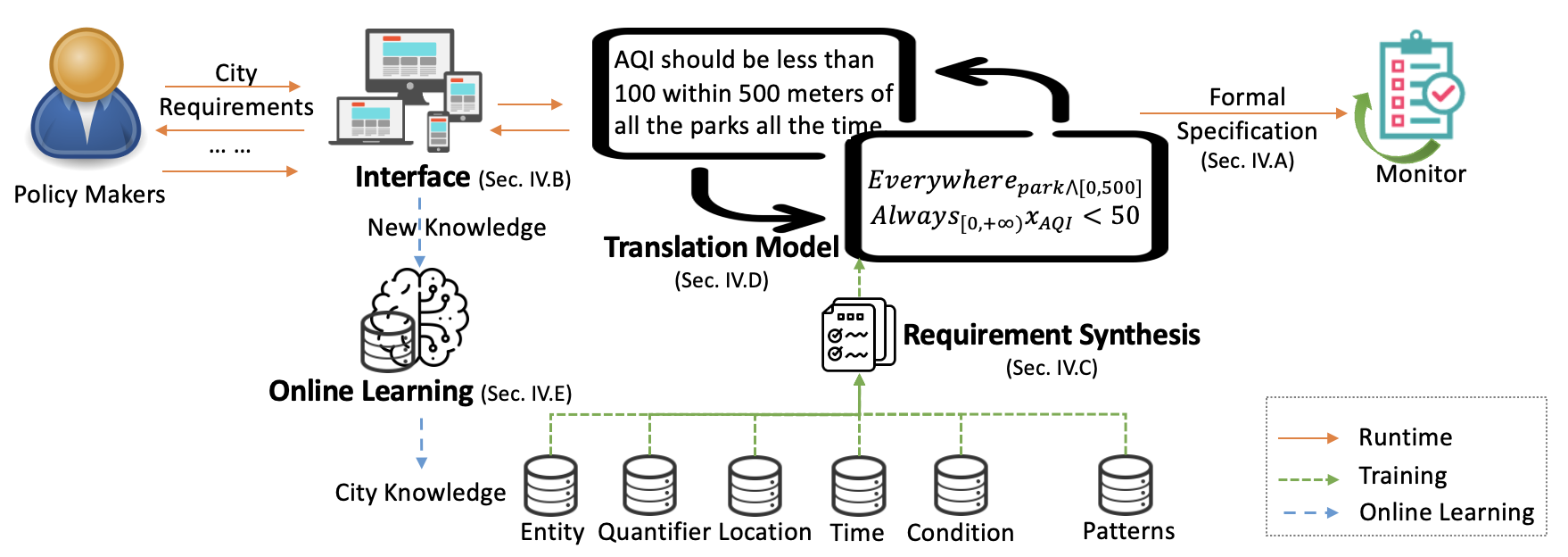}
    \caption{System Overview}
    \label{fig:Overview}
    \vspace{-0.5cm}
\end{figure*}

CitySpec is designed to bridge the gap between city policy makers and monitoring systems. It supports policy makers to precisely write city requirements in English through an intelligent interface, and then converts them to formal specifications automatically. An overview of CitySpec is shown in Figure \ref{fig:Overview}. There are four major components in CitySpec, including an intelligent assistant \textit{Interface} to communicate with policy makers (see Section \ref{subsec:interface}), a \textit{Requirement Synthesis} component to extract city knowledge and synthesize new requirements to build the translation model (see Section \ref{subsec:requirement}), a \textit{Translation Model} to convert city requirements to formal specifications (see Section \ref{subsec:translation}),  and an \textit{Online Learning} component to adapt the system to new knowledge (see Section \ref{subsec:online}). 

At runtime (as indicated by the orange arrows in Figure \ref{fig:Overview}), 
a city policy maker inputs a requirement in English through an intelligent assistant {interface}, which sends the requirements to the translation model. The \textit{translation model} converts the requirements to a formal specification and checks if there is any missing information or ambiguous description. 
The translation model is built with injected city knowledge through requirement synthesis at the training time and enhanced through online learning at runtime. 
Next, based on the returned results from the translation model, the intelligent interface communicates with the policy maker to acquire or clarify the essential information. In this process, the assistant supports the policy maker to refine the requirement until it is precisely defined and accepted by the monitor. We present the technical details in Section \ref{sec:method}, 
and develop a prototype tool of the CitySpec system and deploy it online.

\section{Methodology}
\label{sec:method}
In this section, we present the major components in CitySpec (as shown in Figure \ref{fig:Overview}). We first introduce requirement specification using Spatial-aggregation Signal Temporal Logic (SaSTL)~\cite{ma2021novel}.  Then we show the design and technical details of the intelligent assistant interface, requirement synthesis, translation model, and online learning, respectively.

\subsection{Requirement Specification using SaSTL}
% \subsection{SaSTL Syntax}

SaSTL is a powerful formal specification language for Cyber-Physical Systems. We select it as our specification language because of its advantages of expressiveness and monitoring for smart cities. However, CitySpec is general and can work with other specification languages.
SaSTL is defined on a multi-dimensional \emph{spatial-temporal signal} as $\omega: \mathbb{T} \times L \to \{{\mathbb{R}\cup\{\bot\}\}} ^n$,
where $\mathbb{T}=\mathbb{R}_{\ge 0}$, represents the continuous time and $L$ is the set of locations. $X= \{x_1, \cdots, x_n \}$ is denoted by the set of 
variables for each location. 
The spatial domain $\ra$ is defined as, $\ra :=  ([d_1, d_2],\psi)$, $\psi := \top\;|\;p\;|\;\neg\;\psi\;|\;\psi\;\vee\;\psi$,
% \begin{equation*}
% \begin{array}{cl}
%      \ra :=&  ([d_1, d_2],\psi) \\
%      \psi :=& \top\;|\;p\;|\;\neg\;\psi\;|\;\psi\;\vee\;\psi 
% \end{array}
% \end{equation*}
where $[d_1, d_2]$ defines a spatial interval with $d_1 < d_2$ and $d_1,d_2 \in \mathbb{R}$, and $\psi$ specifies the property 
over the set of propositions
that must hold in each location.

The syntax of SaSTL is given by 
\begin{equation*} 
\begin{array}{cl}
\varphi  :=& x \sim c\;| \
 \neg \varphi \;| \
\varphi_1 \land \varphi_2 \;|\
\varphi_1 \until \varphi_2 \;|\
\ag^{\op} x \sim c \;|\
\ct^{\op} \varphi \sim c \\
\end{array}
\end{equation*}

where $x \in X$, $\sim \in \{<, \le\}$, $c \in \mathbb{R}$ is a constant, 
$I \subseteq \mathbb{R}_{> 0}$ is a real positive dense time interval, 
$\until$ is the \emph{bounded until} temporal operators from STL. 
The \emph{always} (denoted $\always$) and \emph{eventually} (denoted $\eventually$) temporal operators can be derived the same way as in STL, where $\eventually \varphi \equiv \mathsf{true} \ \until \varphi$, and $\always \varphi \equiv \neg \eventually \neg \varphi$.
Spatial \emph{aggregation} operators $\ag^{\op} x \sim c$ for 
$\op \in \{\max, \min, \mathrm{sum}, \avg\}$ evaluates the aggregated product of traces $\op(\nbx)$ over a set of locations $l \in \nb^l$, and \emph{counting} operators $\ct^{\op} \varphi \sim c$ for 
$\op \in \{\max, \min, \mathrm{sum}, \avg\}$ counts the satisfaction of traces over a set of locations. 
From \textit{counting} operators, we derive the \emph{everywhere} operator as $\ew \varphi \equiv \ct^{\mathrm{min}} \varphi > 0$, and \emph{somewhere} operator as $\sw \varphi \equiv \ct^{\mathrm{max}} \varphi > 0$. Please refer to \cite{ma2021novel} for the detailed definition and semantics of SaSTL.

\subsection{Interface for Intelligent Assistant}
\label{subsec:interface}

City requirements often have missing or ambiguous information, which may be unnoticed by policy makers. It leads to the demand for human inputs and clarification when converting them into 
formal specifications. Therefore, we design an intelligent assistant interface in CitySpec serving as an intermediary between policy makers and the translation model. It communicates with policy makers and confirms the final requirements through an intelligent conversation interface. 

To briefly describe the communication process, users first input a requirement in English, e.g., ``due to safety concerns, the number of taxis should be less than 10 between 7 am to 8 am''. CitySpec interface passes the requirement to the translation model and gets a formal requirement ($\mathsf{always}_{[7,8]} \mathsf{number~of~taxi}<10$) with the keywords including,  
\begin{itemize}
    \item $\mathsf{entity}$: the requirement's main object, e.g., ``the number'',
    \item $\mathsf{quantifier}$: the scope of an entity, e.g., ``taxi'',
    \item $\mathsf{location}$: the location where this requirement is in effect, which is missing from the above example requirement, 
    \item $\mathsf{time}$: the time period during which this requirement is in effect, e.g., ``between 7 am to 8 am'', 
    \item $\mathsf{condition}$: the specific constraint on the entity, such as an upper or lower bound of $\mathsf{entity}$, e.g., ``10''.  
\end{itemize}

As a result, CitySpec detects that the location information is missing from the user's requirement and generates a query for the user, ``what is the location for this requirement?'' Next, with new information typed in by the user (e.g., ``within 200 meters of all the schools''), CitySpec obtains a complete requirement. 

The next challenge is how to confirm the formal specification with policy makers. Since they do not understand the formal equation, we further convert it to a template-based sentence. Therefore, CitySpec presents three formats of this requirements for users to verify, (1) a template-based requirement, e.g., [number] of [taxi] should be [$<$] [10] [between 7:00 to 8:00] [within 200 meters of all the schools], (2) a SaSTL formula $\mathsf{everywhere}_{\mathsf{school}\land[0,200]}\mathsf{always}_{[7,8]} \mathsf{number~of~taxi}<10$, and (3) five key fields detected. Users can confirm or further revise this requirement through the intelligent assistant. 

When policy makers have a large number of requirements to convert, to minimize user labor to input requirements manually, CitySpec also provides the option for them to input requirements through a file. The process is similar to where CitySpec asks users to provide or clarify information until all the requirements are successfully converted through files.

% structure of method: 
% background* in order to xxxx 
% what is it, regarding the whole system; novel? 
% why do we need it/ goal/ purpose 
% input and output; 
% each key component or tech (and why using them)

\subsection{Requirement Synthesis}
\label{subsec:requirement}
% Tackle the challenge of 
The amount of city requirement dataset is insufficient to train a decent translation model in an end-to-end manner. As we've discussed in Section \ref{sec:introduction}, it requires extensive domain knowledge in both city and formal specifications and is extremely time-consuming to annotate new requirements. Furthermore, a majority of the existing city requirements are qualitatively or imprecisely written, which cannot be added to the requirement dataset without refinement~\cite{ma2021novel}. 
To mitigate the challenge of small data to build a translation model, we design a novel approach to incorporating city knowledge through controllable requirement synthesis. 
% It enriches the dataset with both semantic-based and syntactic-based features. 
 
% challenges: 
% vocabulary
% pattern 

There are two main reasons why converting a city requirement to a formal specification is challenging with a small amount of data. First, \textit{the vocabulary of city requirements are very diverse}. For example, requirements from different cities (e.g., Seattle and New York City) or in different domains (e.g., transportation and environment) have totally different vocabulary for entities, locations, and conditions. Second, \textit{the sentence structure (patterns) of requirements vary significantly when written by different people.} It is natural for human beings to describe the same thing using sentences.  

Targeting these two challenges, we first extract city knowledge and build two knowledge datasets, i.e., a vocabulary set and a pattern set. The vocabulary set includes five keys of a requirement, i.e., entity, quantifier, location, time and condition. The pattern set includes requirement sentences with 5 keywords replaced by their labels. For instance, we have a requirement, ``In all \textit{buildings}/$\mathsf{location}$, the average \textit{concentration}/$\mathsf{entity}$ of \textit{TVOC}/$\mathsf{quantifier}$ should be no more than \textit{0.6 mg/m3}/$\mathsf{condition}$ for \textit{every day}/$\mathsf{time}$.'', the pattern extracted is ``In \#$\mathsf{location}$, the average \#$\mathsf{entity}$ of \#$\mathsf{quantifier}$ should be no more than \#$\mathsf{condition}$ for \#$\mathsf{time}$.'' 

We extract the knowledge set from city documents besides requirements so that we are not limited by the rules of requirements and enrich the knowledge of our model.  
For example, we extract 336 patterns and 3061 phrases (530 phrases in $\mathsf{entity}$, 567 phrases in $\mathsf{quantifier}$, 501 phrases in $\mathsf{location}$, 595 phrases in $\mathsf{condition}$, and 868 phrases in $\mathsf{time}$). 

% \subsubsection{Pattern v.s. Vocabulary}
% The term vocabulary refers to each predefined named entity. The term pattern is defined to be similar to initial sentences, but all those labeled keywords are replaced by their corresponding labels. 

% For instance, we have the following two sentences in the dataset: (1) In all \textit{buildings}/$\mathsf{location}$, the average \textit{concentration}/$\mathsf{entity}$ of \textit{TVOC}/$\mathsf{quantifier}$ should be no more than \textit{0.6 mg/m3}/$\mathsf{condition}$ for \textit{every day}/$\mathsf{time}$. (2) The \textit{toluene}/$\mathsf{entity}$ should be less than \textit{0.2 mg per m3}/$\mathsf{condition}$ within an hour nearby all \textit{park}/$\mathsf{location}$.

% The patterns in those two sentences are stored and further maintained as follows (``\#'' is designed to be a dummy symbol for future detection and location): (1) In all \#$\mathsf{location}$, the average \#$\mathsf{entity}$ of \#$\mathsf{quantifier}$ should be no more than \#$\mathsf{condition}$ for \#$\mathsf{time}$. (2) The \#$\mathsf{entity}$ should be less than \#$\mathsf{condition}$ within an hour nearby all \#$\mathsf{location}$.

% \subsubsection{Controllable Requirement Synthesis}

Next, we designed an approach to synthesizing controllable requirement dataset efficiently.
Intuitively, we can go through all the combinations of keywords and patterns to create the dataset of requirements, which is infeasible and may cause the model overfitting to the injected knowledge. 
% However, as we can have multiple keywords in the pattern, include all the combinations of keywords in the dataset is infeasible. 
% To handle this problem, we propose an efficient algorithm that synthesize the appropriate number of requirements for the model training. 
In order to enhance the model's performance, we need to keep a balance between the coverage of each keyword and the times of keywords being seen in the generation. 
We denote $\lambda$ as the synthesis index, which indicates the \emph{minimum} number of times that a keyword appears in the generated set of requirements. Assuming we have $m$ set of keywords vocabularies $\{V_1, V_2 \dots V_m\}$ and a pattern set as $P$, we have the total number of synthesized requirements $\ell = \lambda \cdot \max(|V_1|, |V_2|, \dots, |V_m|)$. For each set of vocabularies $V_i$, we first create a random permutation of $V_i$ and repeat it until the total number of phrases reaches $\ell$, then we concatenate them to an array $S_i$. Once we obtain $S_1, ... S_m$, we combine them with pattern $P$ to generate a requirement set $R$. Refer to Algorithm \ref{alg:data_enrich} for more details.

\begin{algorithm}[t]
 \caption{Requirement Synthesis}
 \label{alg:data_enrich}
 \scriptsize
\textbf{Input:} $m$ set of keywords vocabularies $\{V_1, V_2 \dots V_m\}$, Pattern $P$, synthesis index $\lambda$ \\
\textbf{Output:} Set of requirements $R$
\begin{algorithmic}
\State Initialize $R$ as an empty set$\{\}$
\State Let $\ell = \lambda \cdot \max(|V_1|, |V_2|, \dots, |V_m|)$
\For{$i \in 1 \dots m$}
\State Initialize $S_i$ as an empty array $S_i = []$
\While{$|S_i| < \ell$}
    \State Create a random permutation of $V_i$: $P = \text{Permutate}(V_i)$ 
    \State Concatenate $P$ to $S_i$: $S_i = \text{Concat}(S_i, P)$
\EndWhile
\EndFor
\For{$j \in 1 \dots \ell$}
    \State Combine keywords $S_1[j], S_2[j] \dots S_m[j]$ with Pattern $P$ to create a requirement $r_j$
    \State Add $r_j$ to the set of requirements $R$
\EndFor
\State \Return $R$
\end{algorithmic}
\end{algorithm}

\subsection{Translation Model}
\label{subsec:translation}

The inputs of the translation model are requirements, and the outputs of this module are formal specifications with token-level classification. 
We implement the translation model with three major components, a learning model, knowledge injection through synthesized requirements, and keyword refinement. 

To be noted, CitySpec does not build its own translation model from scratch. Instead, we tackle the limitation of the traditional language model and improve it for city requirement translation. Therefore, CitySpec is compatible with different language models. 

In this paper, we implement four popular language models, which are Vanilla Seq2Seq, Stanford NLP NER, Bidirectional Long Short Term Memory (Bi-LSTM) + Conditional Random Field (CRF) and Bidirectional Encoder Representations from Transformers (BERT)~\cite{devlin2018bert}. 
We apply our synthesized datasets with different synthesis indexes to inject city knowledge into these language models.  
Then we evaluate the improvement brought by our requirement synthesis approach by testing the performance on real-world city requirements. We present the detailed results and analysis in Section \ref{sec:eval}. 

Additionally, we find that \textit{time, negation and comparison} are the most tricky elements that affect the accuracy of the final specification detection. Therefore, we implement another refinement component in the translation model.   
% \textit{Time refinements:}
In general, the $\mathsf{time}$ domain can be represented in several formats, such as timestamps, or other formats like yyyy-mm-dd and mm-dd-yyyy. To mitigate the confusion that various formats might bring, we apply SUTime~\cite{chang2012sutime} when the $\mathsf{time}$ entity is not given by the translation model. 
% However, if the $\mathsf{time}$ entity has already been fulfilled in the answer-sheet, the answers from SUTime will be neglected. 
% \textit{Negation and comparison refinements: }
% A real mathematical representation of the comparison symbol in input sentences is needed to generate formal specifications. To detect comparison in the sentence, a template-based keyword mapping is adopted. Further,
pyContextNLP~\cite{chapman2011document} is applied to analyze whether there is a negation in the input sentence. If there is any negation, the comparison symbol is reversed. For instance, if there is a keyword ``greater than'', the comparison symbol is $>$. However, if the whole phrase is ``is not supposed to be greater than'', and a negation is detected, thus the final comparison is $\leq$ instead.

\subsection{Online Learning}
\label{subsec:online}
In general, the more clarifications are needed from the users, the worse experience the users will have, especially if users have to clarify the same information repeatedly. For example, if a user from a new city inputs a location that the system fails to detect, the user will be asked to clarify the location information. The user's experience will drop if the system asks him again on the second or third time seeing these words. However, the deep learning-based translation model cannot ``remember'' this information at deployment time. Thus, the first question is that \textit{how can CitySpec learn the new knowledge online? }

Meanwhile, the new information provided by users may also harm the system if it is an incorrect or adversarial example. The second question is that \textit{how can CitySpec validate the new knowledge before learning it permanently?}   

\begin{figure}[t]
    \centering
    \includegraphics[width=0.48\textwidth]{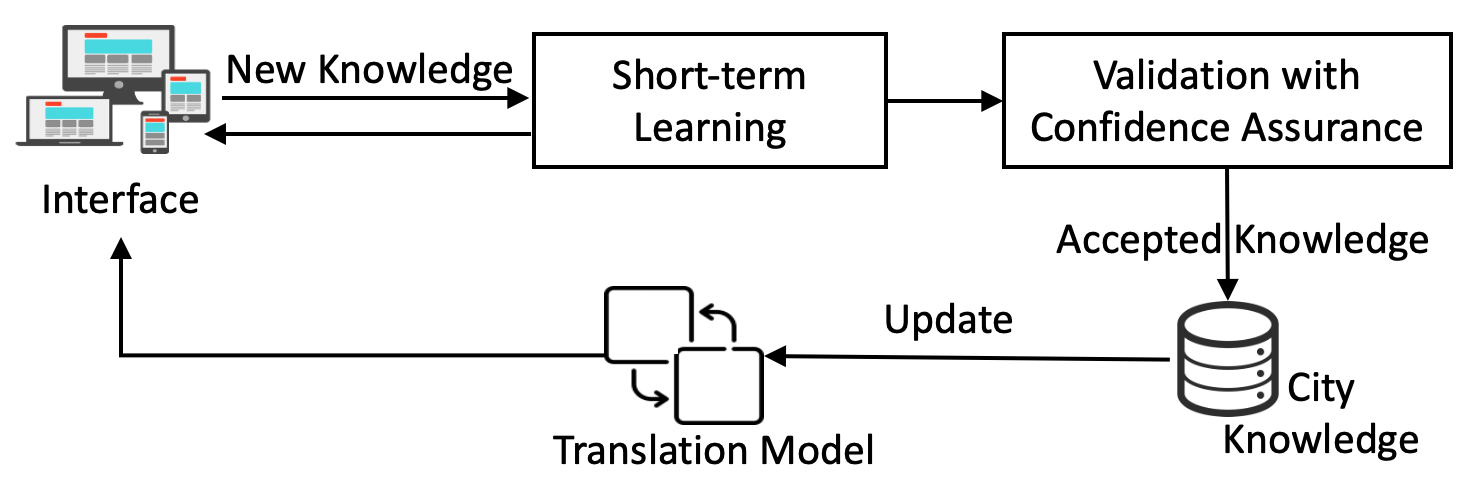}
    \caption{Online Learning}
    \label{fig:online}
    \vspace{-0.8cm}
\end{figure}

Targeting these two research questions, we design an online learning module in CitySpec. As shown in Figure \ref{fig:online}, it has two stages, which are short-term learning and long-term learning. Short-term learning is designed to accommodate the same user in one session of requirement specification with a temporary memory. The question-answer pairs are stored temporarily. When the same occasion occurs, the temporal cache gives instant answers and avoids more user clarifications. 
Long-term learning is designed to adapt the new knowledge to the model permanently after validating its reliability. 
The accepted permanent knowledge is achieved by updating weights via back-propagation on the extended dataset with both initial data and the new input-label pairs stored in the temporary cache. 

To keep CitySpec away from the adversarial inputs, we develop a Bayesian CNN-based validation module in CitySpec. The model is to classify the category of a new term with confidence with uncertainty estimation. We apply dropout layers during both training and testing to quantify the model uncertainty~\cite{ma2021predictive}.  
The inputs of the validation model are the new terms provided by the user, while the outputs are the corresponding keys among those five key domains with an uncertainty level. 
In brief, a new term-key pair is rejected if (1) the output from the validation function does not align with the given domain key; (2) the validation function has low confidence in the output although it might align with the given domain. In this way, we only accept new city knowledge validated with high confidence.

\section{Evaluation }
\label{sec:eval}

In this section, we evaluate our CitySpec system from five aspects, including (1) comparing different language models on the initial dataset without synthesizing, (2) analyzing the effectiveness of the synthesized requirements by enhancing the models with city knowledge, (3) evaluating the performance of the online validation model, (4) testing CitySpec's adaptability in different cities and domains, and (5) an overall case study. We use the city requirement {dataset} described in Section \ref{sec:motivation}. 
To evaluate the prediction of keywords and mitigate the influence caused by different lengths requirements, 
we choose to use \textit{token-level accuracy (token-acc)} and \textit{sentence-level accuracy (sent-acc)} as our main {metrics}. The token-level accuracy aims to count the number of key tokens that are correctly predicted. The sentence-level accuracy counts the prediction as correct only when the whole requirement is correctly translated to a formal specification using SaSTL. Thus, sentence-level accuracy serves as a very strict criterion to evaluate the model performance. 
We also provide the results using other common metrics including precision, recall, and F-1 score. 
The experiments were run on a machine with 2.50GHz CPU,
32GB memory, and Nvidia GeForce RTX 3080Ti GPU.

\subsection{Performance of language models on the initial dataset}

% what do you want to show: models fail to perform well on initial dataset (how to measure 'well') -> we need data augmentation
% what did you do exactly: test model performance on initial dataset in terms of token-level acc, sent-level acc, f1
% what results (table, data, figure): table III
% what observation: seq2seq and stanford tagger are not suitable in this task (explain the failure; these two models will not be discussed in further analysis)
\begin{table*}[t]
    \centering
        \caption{Performance of language models on the initial dataset}
        \scriptsize
 \begin{tabular}{||c | c | c| c | c | c ||} 
 \hline
 \textbf{Model} & \textbf{Token-Acc} & \textbf{Sent-Acc} & \textbf{F-1 Score} & \textbf{Precision} & \textbf{Recall} \\ [0.2ex] 
 \hline\hline
 \textbf{Vanilla Seq2Seq} & \ 10.91 $\pm$ 0.57 \% & \ 1.38 $\pm$ 0.49\% & \ 24.12 $\pm$ 0.24 \% & \ 65.58 $\pm$ 7.95 \% & \ 14.81 $\pm$ 1.58 \%\\ 
 \hline
 % Pretrained Stanford NER & \ 1.23 \% & \ 0 \% & \ 0 \% & \ 0 \% & \ 0 \%\\ \hline
\textbf{BiLSTM + CRF} & \ 77.59 $\pm$ 0.52 \% & \ 60.82 $\pm$ 1.22 \% & \ 80.46 $\pm$ 0.84 \% & \ 81.11 $\pm$ 1.38 \% & \ 79.83 $\pm$ 7.24 \%\\
 \hline
%  BiLSTM + CRF with GloVe(t) & \ 83.72 $\pm$ 0.84\% & \ 50.21 $\pm$ 2.70 \% & \ 79.15 $\pm$ 1.16 \% & \ 82.59 $\pm$ 1.54 \% & \ 76.04 $\pm$ 2.20 \%\\
%  \hline
 \textbf{BERT} & \ 80.41 $\pm$ 0.07 \% & \ 59.02 $\pm$ 0.42 \% & \ 81.43 $\pm$ 0.01 \% & \ 78.62 $\pm$ 0.01 \% & \ 84.46 $\pm$ 0.01 \%\\
 \hline
\end{tabular}
    \label{tab:initial}
    \vspace{-0.3cm}
\end{table*}

As a baseline of translation model without city knowledge,  
we first evaluate the performance of CitySpec using different language models, including Vanilla Seq2Seq, pretrained Stanford NER Tagger, Bi-LSTM + CRF, and BERT on the initial dataset. We present the results in Table \ref{tab:initial}. 

We make the following observations from the results. 
First, the overlap between Stanford Pretrained NER Tagger prediction and vocabulary is only 9 out of 729. The pretrained tagger tends to give locations in higher granularity. Since this task is a city-level, more detailed location information is stated in a lower granularity by providing the street name, building name, or community name. The $\mathsf{location}$ domain in the pretrained tagger gives more high-level information like city name, state name, or country name. For example, ``34th Ave in Nashville, the state of Tennessee'' is annotated as $\mathsf{location}$ in this task, however, the pretrained NER tagger gives ``Tennessee" as $\mathsf{location}$ instead. 
% Therefore, we leave out the evaluation on it in the rest of this section. 

Secondly, the testing token-acc from Vanilla Seq2Seq is 10.91\% on average. Other metrics also indicate that Vanilla Seq2Seq has trouble recognizing the patterns in sequential keyword labeling. 
The Vanilla Seq2Seq model suffers from data scarcity and has difficulty recognizing the general patterns in the training samples due to the small size of the dataset.

Thirdly, the Bi-LSTM + CRF and BERT model achieve better performance than other models, and BERT models often outperform other models with lower standard deviation. 
However, the highest token-level accuracy achieved is 80.41\%, which is still not high enough for an accuracy-prioritized task. A different key may change the requirement entirely. For example, the ``width'' of ``car windshield'' and the ``width'' of ``car'' focus on completely different aspects, although the keywords ``car windshield'' and ``car'' have an only one-word discrepancy. Meanwhile, the best sentence-level accuracy achieved is 60.82\%, which means that about 40\% of the requirements are falsely translated. Assuming the policy makers fix these requirements through the intelligent assistant interface, it is time-consuming and reduces the user experience. Even worse, it may bring safety issues to the monitoring system without noticing.

% ==============
% These requirements will either be fixed by the policy makers and lead to bad user experience or remain unnoticed and cause safety issues in the future. 

% are still not high enough for an accuracy-prioritized task

% Although the highest token-level accuracy achieved is 80.41\%, it is still not high enough for an accuracy-prioritized task, since a different key will result in a different consequence. For example, the ``width'' of ``car windshield'' and the ``width'' of ``car'' focus on completely different aspects, although the keyword ``car windshield'' and ``car'' have only one 
% word difference. As a result, the best sent-acc of BiLSTM and BERT are only 60.82\% and 59.02\%, respectively. 

\textit{In summary, the results indicate that existing language models are not sufficient to serve as the translation model for CitySpec directly. There is a high demand for injecting city knowledge to build the translation model. 
}
% The training process suggests a bigger and more general dataset, which provides more guidance to the model inference.

\subsection{Requirement Synthesis with City 
Knowledge}

\begin{figure}[t]
    \centering
    \includegraphics[width=0.5\textwidth]{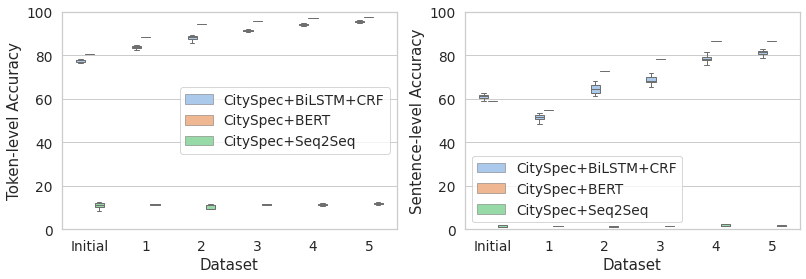}
    \caption{Performance improvement brought by requirement synthesis}
    \label{fig:exp_res}
    \vspace{-0.6cm}
\end{figure}

% what do you want to show: the augmentation algorithm
% what did you do exactly: see algorithm details at alg.1
% what results (table, data, figure): figure 4 and figure 5
% what observation: augmented dataset helps model learn (overall performance and keyword-level performance)

\begin{figure}[t]
    \centering
    \includegraphics[width=0.49\textwidth]{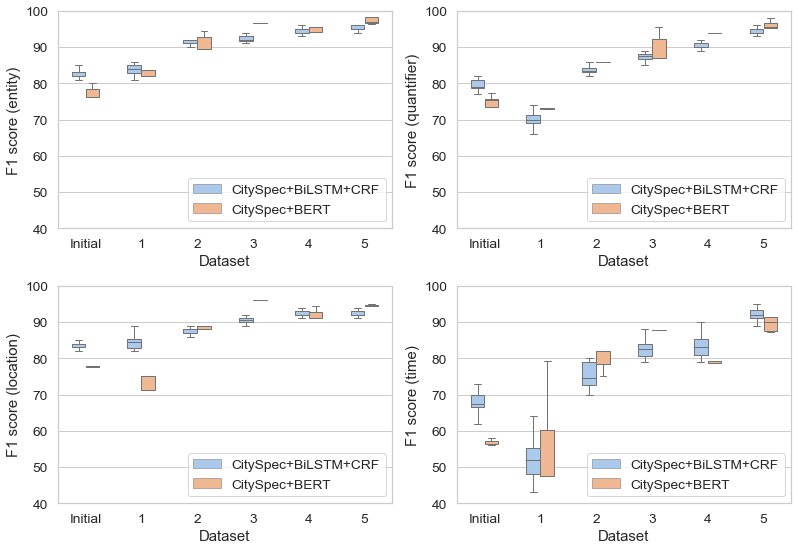}
    \caption{F-1 scores on four keywords}
    \vspace{-0.5cm}
    \label{fig:key_res}
\end{figure}

% In our experiment settings, each experiment configuration is run 20 times, 5 times for the BERT models, independently. 
Next, we evaluate CitySpec's performance with our controllable synthesized requirements. For a fair comparison, we do not use the requirements in the testing set to synthesize requirements. We ensure that the trained model has not seen the requirements in the testing set in either the knowledge injection or training phases.  
% In requirement synthesis, the dataset is generated by initial patterns and combined vocabulary which contains initial vocabulary from over 1,500 requirements. 
% The size of dataset equals to the product of the index and the size of the initial dataset. 
% The evaluation results are represented with mean and standard deviation correspondingly. 
We apply different synthesis indexes to test the effects on the prediction performance. We present the overall results on token-level and sentence-level accuracy in Figure \ref{fig:exp_res}, and F-1 scores on individual keyword in Figure \ref{fig:key_res}. In the figures, x-axis represents the synthesis index. When the index equals to ``inital'', it shows the results without synthesis data.

From the results, we find that, for BERT and Bi-LSTM, there is an overall increase in performance in all token-level accuracy, sentence-level accuracy, overall F-1 score, and F-1 score on keywords. For example, BiLSTM+CRF's token-level accuracy increases from 77.59\% to 97\% and sentence-level accuracy increases from 60.82\% to 81.3\%, BERT's sentence-level accuracy increases from 59.02\% to 86.64\%. 
% Meanwhile, it did not result in any obvious improvement for the seq2seq model. 
% Secondly, there is a slight downgrade when index equals to 1, which is caused by the surging of new vocabulary brought by new introduced cities. 

\textit{In summary, the results show that injecting city knowledge with synthesized requirements boosts the translation model significantly. While improving policy maker's user experience with higher accuracy and less clarifications, it also enhances the safety of the monitoring system potentially.}

% This leap shows the advantages of injecting city knowledge to CitySpec while embedded with models like BiLSTM and BERT.

\subsection{Performance on Online Validation }

% what do you want to show: the validation function helps filter out malicious inputs but accept valid input at the same time
% what did you do exactly: set up rejection and acceptation pipeline (direct and indirect conflicts); explain the online learning workflow in a high level(retraining models)
% what results (table, data, figure): rejection rate; acceptance rate of four different scenarios
% what observation: actual numbers of the rejection and acceptance rates

We evaluate the validation model through simulating four different testing scenarios: (I) randomly generated malicious input based on the permutation of letters and symbols; (II) all street names in Nashville; (III) real city vocabulary generated from Nashville requirements; (IV) generated float numbers with different units. 

First of all, the accuracy of validation model is very high. When the uncertainty threshold is set to 0.5, i.e., all inputs cause an uncertainty higher than 0.5 will be ruled out, CitySpec gives 100\% success rate against scenario I among 2,000 malicious inputs, 91.40\% acceptance rate among 2,107 samples in scenario II, 92.12\% acceptance rate among 596 samples in scenario III, and 94.51\% acceptance rate among 2,040 samples in scenario IV. 
Additionally, we find that the validation function easily confuses $\mathsf{entity}$ with $\mathsf{quantifier}$ if no further guidance is offered. We look into dataset and figure out $\mathsf{entity}$ and $\mathsf{quantifier}$ are confusing to even humans without any context information. Take the requirement ``In all buildings, the average concentration of Sulfur dioxide (SO2) should be no more than 0.15 mg/m3 for every day.'' As an example, $\mathsf{entity}$ is ``concentration'' and $\mathsf{quantifier}$ is ``Sulfur dioxide (SO2)''. If the requirement is changed to ``The maximum level of the concentration of Sulfur dioxide (SO2) should be no more than 0.15 mg / m3 for every day.'', then $\mathsf{entity}$ is ``maximum level'' and $\mathsf{quantifier}$ is ``concentration'' instead. In addition, terms like ``occupancy of a shopping mall'', ``noise level at a shopping mall'', and ``the shopping mall of the commercial district'' also introduce confusion between $\mathsf{location}$, $\mathsf{entity}$ and $\mathsf{quantifier}$, since the same token ``shopping mall'' can be $\mathsf{entity}$, $\mathsf{quantifier}$ or $\mathsf{location}$ in certain cases. 
% Similar cases can happen in $\mathsf{condition}$ and $\mathsf{time}$ domains. 

\textit{The results show that the validation algorithm can effectively accept new city knowledge, prevent adversarial inputs and safeguard online learning. Therefore, CitySpec reduces unnecessary interactions between policy makers and the system and increases efficiency.}

% It is obvious to find that same tokens can be categorized into different domains. In CitySpec Guard settings, we merge $\mathsf{location}$, $\mathsf{entity}$, and $\mathsf{quantifier}$ into a generic key, and $\mathsf{condition}$ and $\mathsf{time}$ as another generic key. By ruling out the conflicts between guard predictions and the clarification given by the user, we can first avoid direct conflicts between model prediction and user clarification (Section \ref{subsec:online}).

% After filtering out direct conflicts, we can then rule out indirect conflicts by setting up an uncertainty threshold. The uncertainty is introduced by applying dropout layers in both training and testing time and further calculated by the inconsistencies among multiple times of predictions caused by inputs. 

\subsection{Adaptability to different scenarios}

% what do you want to show: first the model is not performing well enough when faced with unseen cities (but still not very bad); after adapting the performance is improved
% what did you do exactly: vocabulary separation (topic-wise and city-wise vocab); non-adaptive and adaptive training based on cities and topics
% what results (table, data, figure): table IV and V
% what observation: the adaptability of CogSpec on unseen cities and topics brought by online learning (from a system perspective)

\begin{table*}[]
\caption{Adaptability on different cities in terms of token-level accuracy, sent-level accuracy, and overall F-1 score}
\scriptsize
\centering
\resizebox{\textwidth}{!}{%
\begin{tabular}{|c|ccc|ccc|ccc|ccc|}
\hline
\textbf{City}                    & \multicolumn{3}{c|}{\textbf{Seattle}}                                          & \multicolumn{3}{c|}{\textbf{Changsha}}                                         & \multicolumn{3}{c|}{\textbf{Charlottesville}}                                  & \multicolumn{3}{c|}{\textbf{Jacksonville}}                                     \\ [0.2ex]\hline\hline
\textbf{Metrics}                 & \multicolumn{1}{c|}{\textbf{TokenAcc}} & \multicolumn{1}{c|}{\textbf{SentAcc}} & \textbf{F-1} & \multicolumn{1}{c|}{\textbf{TokenAcc}} & \multicolumn{1}{c|}{\textbf{SentAcc}} & \textbf{F-1} & \multicolumn{1}{c|}{\textbf{TokenAcc}} & \multicolumn{1}{c|}{\textbf{SentAcc}} & \textbf{F-1} & \multicolumn{1}{c|}{\textbf{TokenAcc}} & \multicolumn{1}{c|}{\textbf{SentAcc}} & \textbf{F-1} \\ \hline
\textbf{Non-adaptive w/ BiLSTM+CRF}           & \multicolumn{1}{c|}{84.91\%} & \multicolumn{1}{c|}{48.00\%} & 77.60\% & \multicolumn{1}{c|}{86.61\%} & \multicolumn{1}{c|}{61.20\%} & 84.10\% & \multicolumn{1}{c|}{90.00\%} & \multicolumn{1}{c|}{65.62\%} & 86.48\% & \multicolumn{1}{c|}{77.32\%} & \multicolumn{1}{c|}{35.20\%} & 81.54\% \\ 
\textbf{Adapted w/ BiLSTM+CRF} & \multicolumn{1}{c|}{96.05\%} & \multicolumn{1}{c|}{84.80\%} & 93.75\% & \multicolumn{1}{c|}{95.27\%} & \multicolumn{1}{c|}{83.20\%} & 93.57\% & \multicolumn{1}{c|}{96.82\%} & \multicolumn{1}{c|}{89.29\%} & 95.40\% & \multicolumn{1}{c|}{97.35\%} & \multicolumn{1}{c|}{88.80\%} & 96.02\% \\ \hline
\textbf{Non-adaptive w/ BERT }          & \multicolumn{1}{c|}{80.38\%} & \multicolumn{1}{c|}{46.40\%} & 76.80\% & \multicolumn{1}{c|}{86.10\%} & \multicolumn{1}{c|}{58.00\%} & 83.87\% & \multicolumn{1}{c|}{93.44\%} & \multicolumn{1}{c|}{73.21\%} & 90.46\% & \multicolumn{1}{c|}{90.21\%} & \multicolumn{1}{c|}{56.40\%} & 81.60\% \\ 
\textbf{Adapted w/ BERT}       & \multicolumn{1}{c|}{95.10\%} & \multicolumn{1}{c|}{80.70\%} & 90.28\% & \multicolumn{1}{c|}{97.16\%} & \multicolumn{1}{c|}{88.40\%} & 96.88\% & \multicolumn{1}{c|}{97.53\%} & \multicolumn{1}{c|}{87.05\%} & 94.31\% & \multicolumn{1}{c|}{96.72\%} & \multicolumn{1}{c|}{83.20\%} & 92.59\% \\ \hline
\textbf{Knowledge Injected }    & \multicolumn{3}{c|}{31 (2.34\% )}                                           & \multicolumn{3}{c|}{23 (1.73\%)}                                           & \multicolumn{3}{c|}{53 (4.07\%)}                                           & \multicolumn{3}{c|}{22 (1.66\%)}                                           \\ \hline
% Vocab Introduced & \multicolumn{3}{c|}{31}                                                                  & \multicolumn{3}{c|}{23}                                                                  & \multicolumn{3}{c|}{53}                                                                  & \multicolumn{3}{c|}{22}                                                                  \\ \hline
\end{tabular}

}
\label{tab:cities}
\end{table*}

\begin{table*}[]
\caption{Adaptability on different topics in terms of token-level accuracy, sent-level accuracy, and overall F-1 score}
\resizebox{\textwidth}{!}{%
\begin{tabular}{|c|ccc|ccc|ccc|ccc|}
\hline
\textbf{Topic}                   & \multicolumn{3}{c|}{\textbf{Noise Control}}                                                       & \multicolumn{3}{c|}{\textbf{Public Access}}                                                       & \multicolumn{3}{c|}{\textbf{Indoor Air Control}}                                                  & \multicolumn{3}{c|}{\textbf{Security}}                                                            \\ [0.2ex]\hline\hline
\textbf{Metrics}                 & \multicolumn{1}{c|}{\textbf{TokenAcc}} & \multicolumn{1}{c|}{\textbf{SentAcc}} & \textbf{F-1} & \multicolumn{1}{c|}{\textbf{TokenAcc}} & \multicolumn{1}{c|}{\textbf{SentAcc}} & \textbf{F-1} & \multicolumn{1}{c|}{\textbf{TokenAcc}} & \multicolumn{1}{c|}{\textbf{SentAcc}} & \textbf{F-1} & \multicolumn{1}{c|}{\textbf{TokenAcc}} & \multicolumn{1}{c|}{\textbf{SentAcc}} & \textbf{F-1} \\ \hline
\textbf{Non-adaptive w/ BiLSTM+CRF  }         & \multicolumn{1}{c|}{77.82\%}         & \multicolumn{1}{c|}{41.56\%}        & 77.83\%     & \multicolumn{1}{c|}{73.99\%}         & \multicolumn{1}{c|}{44.07\%}        & 74.41\%     & \multicolumn{1}{c|}{81.51\%}         & \multicolumn{1}{c|}{46.80\%}        & 76.22\%     & \multicolumn{1}{c|}{72.11\%}         & \multicolumn{1}{c|}{28.80\%}        & 62.93\%     \\ 
\textbf{Adapted w/ BiLSTM+CRF }& \multicolumn{1}{c|}{95.82\%}         & \multicolumn{1}{c|}{90.68\%}        & 94.46\%     & \multicolumn{1}{c|}{97.68\%}         & \multicolumn{1}{c|}{74.80\%}        & 97.39\%     & \multicolumn{1}{c|}{96.58\%}         & \multicolumn{1}{c|}{80.00\%}        & 87.68\%     & \multicolumn{1}{c|}{94.39\%}         & \multicolumn{1}{c|}{94.37\%}        & 92.34\%     \\ \hline
\textbf{Non-adaptive w/ BERT }          & \multicolumn{1}{c|}{84.15\%} & \multicolumn{1}{c|}{58.62\%} & 81.05\% & \multicolumn{1}{c|}{83.59\%} & \multicolumn{1}{c|}{54.17\%} & 78.93\% & \multicolumn{1}{c|}{78.51\%} & \multicolumn{1}{c|}{31.20\%} & 73.88\% & \multicolumn{1}{c|}{79.31\%} & \multicolumn{1}{c|}{45.60\%} & 77.50\% \\ 
\textbf{Adapted w/ BERT}       & \multicolumn{1}{c|}{98.07\%}         & \multicolumn{1}{c|}{88.31\%}        & 92.98\%     & \multicolumn{1}{c|}{97.43\%}         & \multicolumn{1}{c|}{88.75\%}        & 95.75\%     & \multicolumn{1}{c|}{95.31\%}         & \multicolumn{1}{c|}{74.40\%}        & 93.60\%     & \multicolumn{1}{c|}{95.41\%}         & \multicolumn{1}{c|}{82.80\%}        & 93.95\%     \\ \hline
\textbf{Knowledge Injected}      & \multicolumn{3}{c|}{57 (4.38\%)}                                                              & \multicolumn{3}{c|}{171 (14.37\%)}                                                              & \multicolumn{3}{c|}{92 (7.31\%)}                                                              & \multicolumn{3}{c|}{41 (3.13\%)}                                                              \\ \hline
% Vocab Introduced & \multicolumn{3}{c|}{57}                                                                  & \multicolumn{3}{c|}{171}                                                                 & \multicolumn{3}{c|}{92}                                                                  & \multicolumn{3}{c|}{41}                                                                  \\ \hline
\end{tabular}}
% \vspace{-0.9cm}
\label{tab:topics}
\end{table*}

In this section, we analyze CitySpec's adaptability in different cities and different domains.  
Different cities have different regulation focuses and their city-specific vocabulary. For example, in the city of Nashville, $\mathsf{location}$ names like ``Music Row'', ``Grand Ole Opry'' will probably never appear in any other cities. We select four cities, Seattle, Charlottesville, Jacksonville, and Changsha, with different sizes and from different countries as case studies. We separate the requirements of each mentioned city and extract the city-wise vocabulary based on each city independently. Each of four constructed pairs consists of: vocabulary I, which is extracted from the requirements from one city only, and vocabulary II, which is extracted from the requirements from all the cities but that specific one city. Injected knowledge is measured using the number along with the ratio of how much of the unique vocabulary one city causes. We augment vocabulary II using 5 as the synthesis index and train a model on vocabulary II. As a result, the trained model is isolated from the vocabulary information from that one specific city. Afterward, we test the trained model performance on the generated requirements using vocabulary I. We pick CitySpec with Bi-LSTM + CRF and CitySpec with BERT in this scenario. We employ the validation function to validate all vocab in vocabulary I and pass the validated ones to vocabulary II. After that, we have a validated vocabulary including vocabulary II and validated vocabulary I. The deployed model is fine-tuned based on the validated vocabulary using few-shot learning.

From the results shown in Table \ref{tab:cities}, we observe that (1) although CitySpec immigrates  to a completely unknown city, it is still able to provide satisfying performance, e.g., 84.9\% token-acc and 77.6\% F-1 score in Seattle, but the sent-acc tends to be low. (2) With new knowledge injected, the performance increases significantly, e.g., Sent-Acc for Seattle increases from 48\% to 84.8\% with BiLSTM+CRF, and from 46.4\% to 80.7\% with BERT.

We also explore CitySpec's adaptability to different topics. We choose four topics including noise control, indoor air control, security, and public access. The results also show that (1) even though CitySpec has not seen vocabulary from a totally different topic, it still gives a competitive performance; (2) online learning brings obvious improvements when adaptation is further applied.

\textit{In summary, it indicates the capability of CitySpec in both city and domain adaptation. It can also adapt to new requirements evolving overtime. 
Moreover, with a different set of domain-specified knowledge, CitySpec can be potentially applied to other application domains (e.g., healthcare). }

\subsection{Case Study} 
Due to the absence of a real city policy maker, we emulate the process of using CitySpec by taking the real-world city requirements and assuming that they are input by policy makers. Specifically, this case study shows the iteration of communication between CitySpec and the policy maker to clarify the requirements. 
We emulate this process 20 times with 100 requirements randomly selected from our datasets each time. The results show that the average and maximum rounds of clarification are 0.8 and 4 per requirement, respectively, due to the missing or ambiguous information. Averagely, 28.35\% of requirements require clarification on location. 
For example, ``No vendor should vend after midnight.'', CitySpec asks users to clarify the time range for ``after midnight'' and the location defined for this requirement. 
Overall, CitySpec obtains an average sentence-level accuracy of 90.60\% (with BERT and synthesize index = 5). The case study further proves the effectiveness of CitySpec in city requirement specification.

% , and learns xxx new phrases online in the case study. 
\section{Related Work}
\label{sec:related}

\textbf{Translation Models.} Researchers have developed models to translate the natural language to machine languages in various domains, such as Bash commands~\cite{fu2021transformer}, Seq2SQL~\cite{zhong2017seq2sql}, and Python codes~\cite{chen2021evaluating}. These translation models benefit from enormous datasets. The codex was trained on a 159 GB dataset that contains over 100 billion tokens. WikiSQL, which Seq2SQL was trained on, consists of 80,654 pairs of English-SQL conversions. NL2Bash \cite{fu2021transformer} was trained on approximately 10,000 pairs of natural language tasks and their corresponding bash commands. 
As an under-exploited domain, there is a very limited number of well-defined requirements. Therefore, existing translation models do not apply to our task. This paper develops a data synthesis-based approach to build the translation model. 

% There have already been pieces of previous work trying to solve the conversion from natural languages to machine languages. Codex~\cite{chen2021evaluating} from OpenAI is a generative model trained on publicly available code from GitHub based on GPT~\cite{radford2018improving} and is able to compose python codes based on English sentences. Other than python code, conversions to Bash commands~\cite{fu2021transformer} and SQL~\cite{zhong2017seq2sql} are also discussed previously. However, computer languages like Bash commands and SQL have a relatively fixed syntax compared with formal specifications. Although Codex is trained to generate python code, it is only able to output code in the form of python functions. The conversion from English to formal specifications is considered to be more difficult in terms of language flexibility and symbol complexity. Although previously proposed approaches are reported to have convincing evaluation results, most of them profit from enormous datasets. Codex is trained on a 159 GB dataset which contains over 100 billion tokens. WikiSQL, which Seq2SQL~\cite{zhong2017seq2sql} was trained on, consists of 80,654 pairs of English-SQL conversion. NL2Bash, which~\cite{fu2021transformer} was trained on, includes approximately 10,000 pairs of natural language tasks and their corresponding bash commands. In this work, only around 1,500 requirements are available. 

\textbf{Data Synthesis.} Data synthesis exploits the patterns in study findings and synthesizes variations based on those patterns. Data augmentation is a simple application of data synthesis. Previous augmentation approaches wield tricks like synonym substitution~\cite{kobayashi2018contextual, zhang2015character} and blended approaches~\cite{wei2019eda}. In the smart city scenario, we need new data samples which fit in the smart city context. Therefore, we extract extra knowledge from smart cities and fully exploit semantic and syntactic patterns instead of applying straightforward tricks like chopping, rotating, or zooming. This paper is the first work synthesizing smart-city-specific requirements to the best of our knowledge. 

\textbf{Online Learning.} Online machine learning
% ~\cite{bottou1998online} 
mainly deals with the situation when data comes available to the machine learning model sequentially after being deployed. Similar to continuous learning, online learning aims to give model accumulated knowledge and improve model performance continuously given incoming learning samples~\cite{parisi2019continual, chen2018lifelong}. Some of the existing papers focus on developing sophisticated optimization algorithms~\cite{hazan2016introduction} or exploiting the differences between new and old samples~\cite{sutton2018reinforcement}. However, these papers do not have a mechanism to detect or prevent adversarial samples online. This paper develops a two-stage online learning process with online validation against potential malicious inputs.  

\section{Summary}
\label{sec:summary}

This paper builds an intelligent assistant system, CitySpec, for requirement specification in smart cities. CitySpec bridges the gaps between city policy makers and the monitoring systems. It incorporates city knowledge into the requirement translation model and adapts to new cities and domains through online validation and learning. The evaluation results on real-world city requirement datasets show that CitySpec is able to support policy makers accurately writing and refining their requirements and outperforms the baseline approaches. In future work, we plan to have CitySpec used by real city policy makers, but this is outside the scope of this paper.
% It shows in the evaluation using real-world city requirement datasets. 
% In future work, we plan to deploy CitySpec into our collaborated city control center in Nashville and further improve it based on the experience of city policy makers. 

% In future work, we will put CitySpec into real applications and improve the system based on the real user experience from city policy makers.

% In future work, we will deploy the system in a city center and improve the CitySpec based on user experience by city policy makers.

%  “technical” accuracy/value of the tool based on datasets. Future work it to have the solution used by real city managers, but this is outside the scope of this paper?

\section*{Acknowledgment}
This work was funded, in part, by NSF CNS-1952096. 

\bibliographystyle{IEEEtran}
\bibliography{ref_new}

\end{document}